# Multi-Context Fusion Transformer for Pedestrian Crossing Intention Prediction in Urban Environments

Yuanzhe Li [1*], Hang Zhong [1], Steffen Müller [1]

[1]Chair of Automotive Engineering, Technische Universität Berlin, Berlin, 13355, Germany
[*]yuanzhe.li@campus.tu-berlin.de

**Abstract: Pedestrian crossing intention prediction is essential for autonomous vehicles to improve pedestrian safety and reduce traffic accidents. However, accurate pedestrian intention prediction in urban environments remains challenging due to the variability of pedestrian behavioral patterns and the presence of multiple environmental factors. In this paper, we propose a multi-context fusion Transformer (MFT) that leverages diverse and complementary contextual attributes across four key dimensions, including pedestrian behavior context, environmental context, pedestrian localization context and vehicle motion context, for accurate pedestrian intention prediction. MFT employs a progressive fusion strategy, where mutual intra-context attention enables bidirectional interactions within each context, thereby facilitating intra-context feature integration, with a context token serving as the context-specific representation. This is followed by mutual cross-context attention, which enables bidirectional interactions across context tokens, with the global CLS token serving as a compact multi-context representation. Finally, guided intra-context attention refines context tokens through directed information aggregation within each context, while guided cross-context attention further strengthens the global CLS token by selectively collecting information across contexts, enabling effective and efficient multi-context fusion. Experimental results demonstrate the competitive performance of MFT against state-of-the-art methods, achieving accuracy rates of 74%, 93%, and 90% on the $JAAD_{beh}$, $JAAD_{all}$, and PIE datasets, respectively. Extensive ablation studies are further conducted to demonstrate the effectiveness of the proposed network architecture and the importance of different contexts. Our code is open-source: https://github.com/ZhongHang0307/Multi-Context-Fusion-Transformer.**

## 1. Introduction

With the rapid advancements in artificial intelligence and sensor technologies in recent years, autonomous vehicles (AVs) have been increasingly deployed in urban environments. The rapid deployment of AVs has raised significant safety concerns for vulnerable road users (VRUs), particularly pedestrians, due to the increasing frequency of pedestrian–AVs interactions. Ensuring pedestrian safety requires AVs to have the ability to anticipate pedestrian intentions. Accurate pedestrian crossing intention prediction supplies critical cues, enabling AVs to make safe decisions and perform reliable control [1]. However, predicting pedestrian intentions in urban environments remains challenging, given the variability of pedestrian behavior and the complex influence of multiple environmental factors, such as vehicle states and traffic conditions.

Predicting pedestrian crossing intention has attracted considerable research interest, with numerous studies employing vehicle-mounted camera footage. Representative datasets such as JAAD [2] and PIE [3] provide publicly accessible video data for analyzing pedestrian behavior in urban environments. Initial research efforts primarily relied on unimodal inputs, where researchers attempted to extract visual appearance features from images [2], spatial position features from pedestrian bounding boxes [8], or kinematic features from body pose data [9] for pedestrian intention prediction. For instance, Fang et al. [9] employed convolutional neural network (CNN)-based pose estimation to extract pedestrian skeletons and stacked multi-frame skeleton sequences as input to support vector machine or random forest classifiers for intention prediction. However, such methods often perform poorly in complex urban environments, where pedestrian crossing decisions are influenced by diverse and dynamically changing factors beyond local pedestrian cues, including road layout, traffic infrastructure, surrounding vehicles, and traffic interactions. Consequently, relying solely on pedestrian skeletal information cannot adequately represent the external context underlying crossing behavior, which may reduce prediction accuracy.

Recent studies have explored multimodal fusion strategies for pedestrian intention prediction by integrating heterogeneous yet complementary modalities. In particular, both visual modalities, such as semantic maps [4], optical flow [10], depth maps [32], and motion-related modalities, including vehicle speed [6], and bounding box coordinates [5], have been utilized to capture both pedestrian-level behavior and scene-level context. For example, PCPA [5] employs separate recurrent neural network (RNN) to model non-visual modalities including vehicle speed, bounding boxes, and pose keypoints, while using a 3D CNN to extract spatiotemporal features from the local RGB images. Temporal attention and modality attention are then sequentially applied to assess the contribution of temporal information within each modality and the relative importance of different modalities for intention prediction. In its extended architecture, Global PCPA [4] further integrates global semantic maps to enhance scene-level semantic understanding. In [11], an additional sensor branch is designed to leverage accelerometer and gyroscope data for modelling pedestrians' motion states, which are then synchronously fused with visual information to predict pedestrian crossing intentions. MTC [32] leverages a Transformer to model non-visual modalities and introduces a spatial fusion mechanism for visual modalities with consistent spatial dimensions, thereby facilitating mutual guidance among them.

Although many existing methods adopt an end-to-end paradigm that directly learns from high-dimensional raw modalities such as RGB images, this strategy suffers from several inherent limitations. First, due to the high dimensionality and redundancy of raw modalities, such methods are computationally expensive and often lead to over-parameterized models that are prone to overfitting to training scenarios, thereby undermining robustness and hindering generalization across diverse traffic environments. Furthermore, features derived from raw modalities are often implicit and entangled, making it difficult to interpret their semantic meanings and assess their individual contributions to intention prediction. In addition, the high dimensionality and variability of raw modalities typically require large-scale training data to achieve robust performance, which increases training cost and reduces practicality. Their effectiveness may further degrade substantially in real-world scenarios where the input is noisy, ambiguous, or incomplete. Although some existing studies have incorporated high-level semantic information into pedestrian intention prediction, such as traffic-context cues including traffic lights and crosswalks [39, 40], their contextual modelling remains incomplete, as other important factors, particularly pedestrian behavioral context, are not taken into account. In [41], pedestrian states, vehicle states, and traffic states are jointly incorporated. However, this method relies on traditional machine learning models, such as support vector machines, which are generally less capable of capturing complex interactions among heterogeneous contexts than deep learning-based approaches.

To address the aforementioned challenges, we propose a multi-context fusion Transformer (MFT) for pedestrian crossing intention prediction in urban environments. Rather than relying on raw modalities or limited high-level semantic cues, MFT leverages compact and semantically explicit numerical attributes to comprehensively represent the contextual factors influencing pedestrian crossing decisions. Specifically, these attributes characterize four complementary dimensions: pedestrian behaviors, environmental conditions, pedestrian localization, and vehicle motion dynamics. By jointly modelling the complementarity among these dimensions, MFT enables a more holistic characterization of pedestrian crossing intentions.

The main contributions are summarized as follows:

(1) We propose a multi-context fusion Transformer (MFT) for pedestrian crossing intention prediction, which encodes heterogeneous contextual cues into semantically explicit numerical attributes and integrates them through a Transformer-based fusion framework. Compared with end-to-end methods based on raw modalities, the proposed framework is more lightweight, scalable, and less dependent on the distribution of raw modalities.

(2) A progressive fusion strategy is designed, where mutual intra-context and cross-context attention enable bidirectional information exchange to capture context-specific and multi-context representations, respectively, while guided intra-context and cross-context attention further refine the corresponding context tokens and the global CLS token through directed information aggregation, enabling effective and efficient multi-context fusion.

(3) Extensive experiments on the JAAD and PIE datasets are conducted to validate the performance of the proposed MFT network, while ablation studies further verify the rationality of its architecture and demonstrate that incorporating more contextual cues improves the performance of intention prediction.

## 2. Related works

Pedestrian crossing intention prediction has attracted extensive research attention, with vehicle-mounted video recordings serving as the predominant data source, while controlled virtual-reality experiments have also been employed to investigate pedestrian crossing behaviours and vehicle–pedestrian interaction mechanisms [47, 48]. Public datasets such as JAAD [2] and PIE [3] provide annotated video sequences collected in urban traffic environments, thereby supporting the development of pedestrian intention prediction methods. Early works mainly utilized unimodal inputs, e.g., RGB image [2], pedestrian bounding box [8], or body pose [9], for pedestrian intention prediction. In [2], the authors investigated pedestrian intention prediction based on static images of the upper and lower body regions. In [7], pedestrian position coordinates are extracted via stereo vision using median disparity computation from bounding box regions, which are subsequently processed by a stacked long short-term memory (LSTM) architecture to capture temporal dependencies for intention prediction. However, these methods often perform poorly in complex urban environments, where pedestrian crossing decisions are influenced by multiple contextual factors. For instance, approaches that rely solely on pedestrian images often overlook critical environmental cues, such as crosswalk availability and surrounding vehicle dynamics, both of which play a crucial role in shaping pedestrian behavior.

Subsequent studies have introduced a variety of multimodal fusion methods to improve pedestrian intention prediction by leveraging complementary information from multiple modalities. Different visual modalities (such as semantic maps in [4], and optical flow in [45]) and motion-related modalities (such as bounding box coordinates in [5] and vehicle speed in [6]) have been integrated to capture both pedestrian-specific behaviors and broader environmental contexts. SF-GRU, introduced in [17], employs a stacked RNN framework to gradually fuse five different modalities, including pedestrian appearance, environmental context, body pose keypoints, bounding box information, and vehicle speed, through hidden state stacking. In [4], the Global PCPA framework integrates a CNN–gated recurrent unit (GRU) pipeline for processing visual modalities with GRU-based encoders for non-visual modalities, and further employs a hierarchical attention mechanism that sequentially applies temporal weighting across frames and modality weighting across different feature sources. V-PedCross framework [8] adopts a virtual-to-real distillation strategy, transferring rich multimodal knowledge (e.g., optical flow, RGB images, and bounding boxes) from a teacher network trained on synthetic data to a compact student model that relies solely on bounding boxes, thereby balancing efficiency and predictive accuracy. Its extension, Gated-S2R-PCP [42], further introduces differentiated syn-to-real knowledge transfer and gated knowledge fusion to adaptively exploit heterogeneous modalities with distinct domain gaps for intention prediction. Dual-STGAT [23] employs a dual-level spatio-temporal graph attention architecture to jointly model pedestrian motion and pedestrian–scene interactions.

Given the remarkable success of Transformers in natural language processing and computer vision, Transformers been extensively adopted in autonomous driving tasks, including end-to-end driving [43], motion planning [38], and perception [27]. Owing to their strong capabilities in feature extraction, temporal modelling, and cross-modal interaction learning, Transformers have been increasingly applied to pedestrian intention prediction [46]. For instance, Action-ViT [30] concatenates image frames of visual modalities along the horizontal axis and transforms non-visual inputs into pseudo-images, which are subsequently fed into a pre-trained Vision Transformer (ViT) for temporal representation learning, followed by modality-level attention. TAMformer [31] employs independent Transformer blocks to encode bounding boxes, pose keypoints, and local context images, and introduces learned attention masks for adaptive temporal modelling. Building upon this design, [14] further introduces textual information to provide more explicit semantic context. In [12], Transformer encoders are used to extract non-visual modality features, while a ViT is employed for visual feature extraction, with an additional image enhancement pipeline introduced to improve prediction robustness under adverse weather conditions. IntentFormer [13] adopts a co-learning multimodal Transformer architecture, where three dedicated Transformer encoders jointly learn RGB images, segmentation maps, and trajectory paths under the supervision of a co-learning adaptive composite loss, thereby facilitating coordinated multimodal representation learning. [35] proposes a lightweight cascaded model for intention prediction, which jointly performs bounding box and intention prediction and enhances trajectory features via text descriptions generated by a pre-trained large language model, together with a center-aware classification module. In [25], a global-local context co-learning (GLCCL) framework is proposed to leverage both local and global context during training for collaborative learning, while relying solely on lightweight motion data for efficient real-time inference.

While the aforementioned methods achieve competitive performance in pedestrian intention prediction, they generally adopt an end-to-end paradigm that directly learns from high-dimensional raw modalities, such as RGB images and semantic maps. However, this strategy suffers from several inherent limitations. First, directly modelling high-dimensional raw modalities often incurs substantial computational overhead. Moreover, the redundancy in such inputs may lead to overly complex models that are susceptible to overfitting to specific training scenarios, thereby limiting robustness and generalization across diverse traffic environments. Furthermore, features learned from raw modalities are often implicit and highly entangled, making their semantic meanings and individual contributions to intention prediction difficult to interpret. Moreover, the reliance on high-dimensional and variable inputs typically demands large-scale training data, increasing training cost and limiting practicality. Their performance may also degrade in real-world scenarios where the inputs are noisy, ambiguous, or incomplete.

Although, some existing studies have attempted to incorporate high-level semantic information into pedestrian intention prediction. For example, Faster-PCPNet [37] introduces quasi-polar coordinates to effectively represent high-level pedestrian location information. AugTrEP [39] shows that additional global traffic context, such as crosswalk visibility and traffic light status, can help improve prediction performance under degraded conditions. Similar findings are reported in [40], where the incorporation of traffic awareness data is shown to significantly enhance model performance. However, such methods mainly exploit global traffic context, while fine-grained pedestrian behavioral attributes, such as gaze toward the ego-vehicle and hand gestures signaling right-of-way, are not explicitly considered, despite their potential relevance to intention prediction. In [41], pedestrian, vehicle, and traffic states are jointly incorporated. However, the use of traditional machine learning models, such as support vector machines, limits its ability to capture complex interactions among heterogeneous contexts. This paper considers a more comprehensive set of contextual factors, including pedestrian behaviors, environmental conditions, pedestrian localization, and vehicle motion dynamics. By leveraging semantically explicit numerical contextual attributes that can be derived from raw sensor data, our method constructs a compact yet informative representation for pedestrian intention prediction. Moreover, a progressive fusion mechanism is designed to effectively integrate multiple contextual cues.

## 3. Problem Formulation

We formulate pedestrian crossing intention prediction as a binary classification task. Given a fixed-length video sequence captured by an onboard camera and the corresponding ego-vehicle motion data, the goal is to predict whether the target pedestrian $j$ will cross at a future time instant, denoted by $a_j \in \{0, 1\}$. Specifically, the observation window is defined as a sequence of $N$=16 consecutive frames over 0.5 s, extracted from a 30 FPS video stream, with the last observed frame denoted by $t$. The prediction horizon, corresponding to the time-to-event (TTE), is defined as the interval from the last observed frame $t$ to the event. Here, the event refers to the crossing initiation frame in crossing cases and the last observable frame in non-crossing cases. The TTE ranges from 1 to 2 s (i.e., 30–60 frames).

## 4. Methodology

### 4.1 Input Representation

The MFT network integrates four types of frame-level context to capture complementary information about pedestrian behaviors, environmental conditions, pedestrian localization, and vehicle motion dynamics. An example of these contexts is illustrated in Fig. 1. The four contexts are defined as follows:

1. Pedestrian behavior context $P$: This context employs discrete numerical attributes to represent pedestrian behavioral states and interaction patterns. For the JAAD dataset, $P$ encompasses five behavioral attributes: motion state (standing or walking), gaze state (looking or not looking), head nod (nodding or none), hand gesture (greeting, yielding, right-of-way, and other categories), and motion direction (lateral, longitudinal, or others). For the PIE dataset, the head nod attribute is not provided. Therefore, $P$ has a shape of $N$×5 for JAAD dataset and $N$×4 for PIE dataset. The detailed category definitions and corresponding encodings for JAAD and PIE datasets are provided in Appendix Tables A1 and A2, respectively.

2. Pedestrian localization context $L$: This context represe-

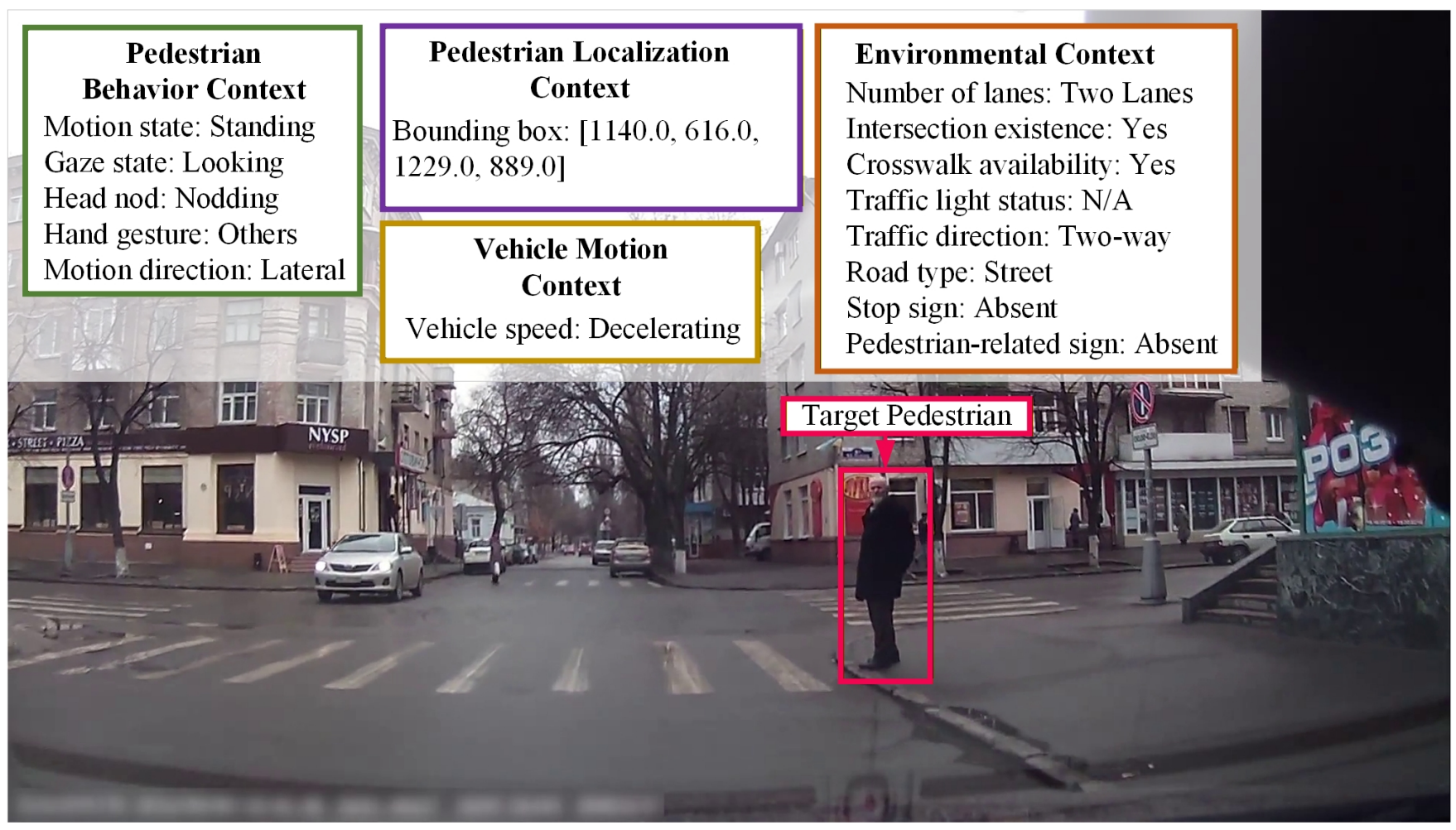

**Fig. 1** Illustration of four contextual annotations in a traffic scene

nts the pedestrian's spatial position in the image using the bounding box coordinates $(x_{tl}^{\tau}, y_{tl}^{\tau}, x_{br}^{\tau}, x_{br}^{\tau})$, where $(x_{tl}^{\tau}, y_{tl}^{\tau})$ and $(x_{br}^{\tau}, x_{br}^{\tau})$ denote the top-left and bottom-right corners at time step $\tau$, respectively, resulting in a input dimension of $N$×4. It captures pedestrian motion and spatial behavior patterns that serve as early cues for crossing intention.

3. Vehicle motion context $V$: This context represents the ego-vehicle's kinematic state and provides important cues for pedestrian intention prediction, as deceleration patterns may indicate yielding behavior and potential crossing opportunities. The vehicle motion context is directly derived from the dataset annotations, without any further preprocessing, such as smoothing, interpolation, or sensor-based estimation. For the PIE dataset, $V$ is represented by the ground-truth ego-vehicle speed values provided in the annotations, whereas for the JAAD dataset, explicit ego-vehicle speed values are not available and $V$ is directly derived from annotated motion states including stopped (0), moving slow (1), moving fast (2), decelerating (3), and accelerating (4). In both datasets, $V$ is represented with an input dimension of $N$×1.

4. Environmental context $E$: This context utilizes discrete numerical attributes to represent the traffic environment. It provides essential information about the road layout, traffic control devices, and other environmental elements that may influence pedestrian–vehicle interactions. For the JAAD dataset, $E$ encompasses eight environmental attributes: number of lanes (single lane, two lanes, three lanes, and other categories), intersection existence (present or absent), crosswalk availability (present or absent), traffic light status (red, green, or N/A), traffic direction (one-way or two-way), road type (street, parking lot, or garage), stop sign presence (present or absent), and pedestrian-related sign presence (present or absent). For the PIE dataset, the road type attribute is not provided. Therefore, $E$ consists of seven environmental attributes and is represented as an $N$×7 input sequence, whereas for the JAAD dataset, $E$ is represented as an $N$×8 input sequence. The detailed definitions of all categories and their corresponding encoding schemes for the JAAD and PIE datasets are provided in Appendix Tables A3 and A4, respectively.

MFT differs from methods based on raw or derived modalities primarily in its input representation. Many existing approaches either directly process raw RGB images or use derived input modalities, such as pose sequences, optical flow, and semantic maps, generated through additional preprocessing or pretrained models. In contrast, MFT uses compact semantic attributes covering pedestrian behavior, environmental context, localization, and vehicle motion. In this study, these attributes are obtained from the JAAD and PIE annotations to enable controlled and reproducible evaluation of the proposed multi-context fusion mechanism. Similar structured behavioral or environmental attributes have also been adopted in recent pedestrian intention prediction studies [40, 41, 44, 49]. In practical intelligent transportation systems (ITS), these attributes can be estimated online by vehicle-side or roadside perception modules. Accordingly, MFT is applicable to vehicle–infrastructure cooperative ITS frameworks, such as the one presented in [44]. Specifically, the pedestrian behavior context $P$, including motion state, gaze state, head nod, hand gesture, and motion direction, can be estimated through pedestrian detection and tracking, followed by pose, gaze, and behavior recognition. The environmental context $E$, encompassing traffic lights, crosswalks, traffic signs, intersections, and road layouts, can be extracted by either onboard perception systems or roadside perception infrastructure with edge-computing capabilities. In the latter setting, the extracted contextual attributes can be disseminated to autonomous vehicles through V2X communication. The pedestrian localization context $L$ can be derived from object tracking, whereas the vehicle motion context $V$ can be acquired directly from onboard sensors.

To represent these contexts in a unified form, the discrete attributes are encoded using label encoding, while the continuous attributes are preserved in their original numerical form. This design facilitates the flexible extension of the framework to additional contexts as well as new attributes within each context. The multi-context fusion Transformer (MFT) adopts a progressive fusion strategy that hierarchically integrates intra-context features and fuses complementary multi-context information for pedestrian intention prediction. The overall architecture of MFT is illustrated in Fig. 2, and detailed descriptions of its key modules are provided in Sections 4.2–4.5.

### 4.2 Intra-Context Fusion

The intra-context fusion (ICF) module performs early-stage fusion within each context to capture temporal dependencies,

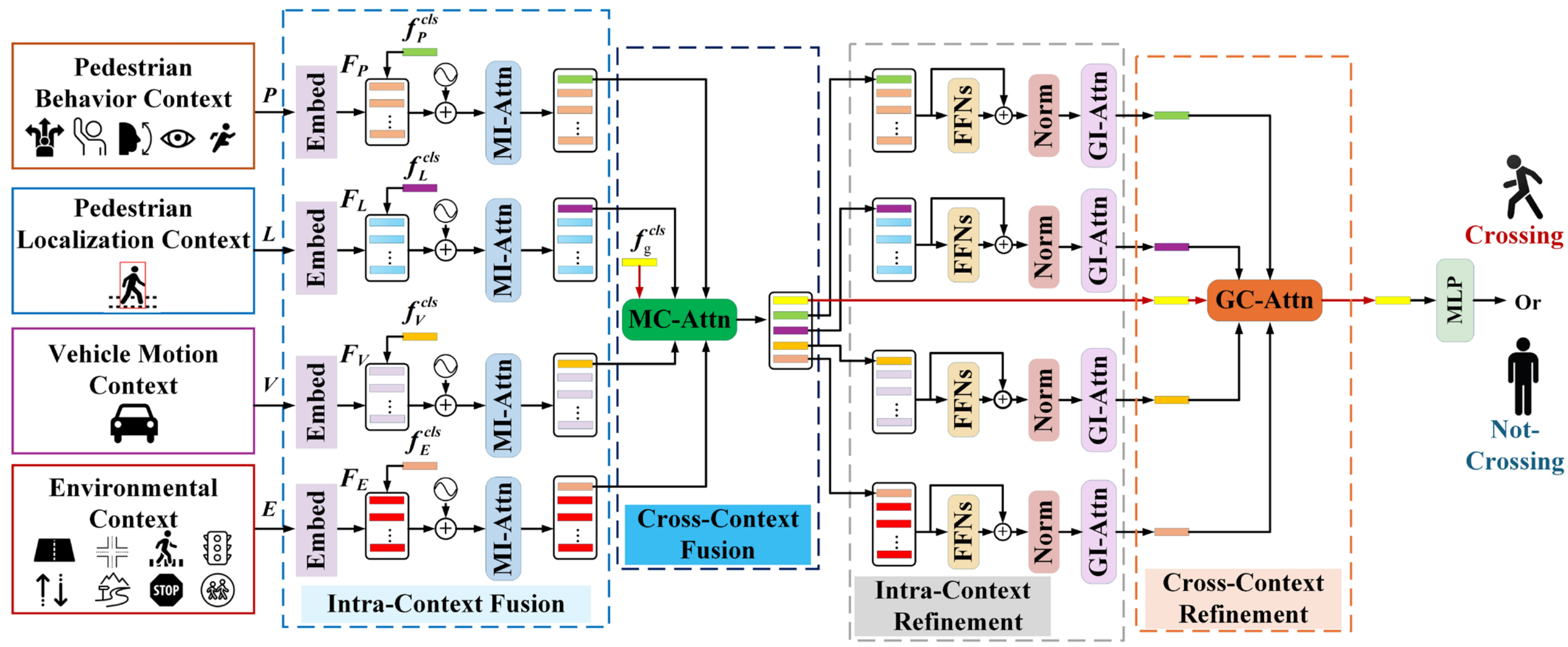

**Fig. 2** The architecture of the proposed MFT network

generating feature sequences with integrated temporal information. Specifically, for context $i$ ($i \in \{P, L, V, E\}$), the raw inputs are first projected to a high-dimensional feature sequence $F_i \in \mathbb{R}^{N\times 128}$ through embedding layers. A learnable context token $f_i^{cls}$ is prepended to the feature sequence $F_i$ along the temporal axis, serving as a compact context-specific representation. Sinusoidal positional encodings (*PE*) are added to maintain temporal order information, producing the position-encoded feature sequence $F_i' \in \mathbb{R}^{(N+1)\times 128}$:

$$F_i' = \left[ f_i^{cls}, f_i^{t-N+1}, \ldots, f_i^{t} \right] + PE \tag{1}$$

where *PE* is defined as:

$$\begin{cases} PE(pos, 2k) = \sin\left(\dfrac{pos}{10000^{\frac{2k}{d}}}\right) \\ PE(pos, 2k+1) = \cos\left(\dfrac{pos}{10000^{\frac{2k}{d}}}\right) \end{cases} \tag{2}$$

where *pos* denotes the position index in the sequence, *k* is the dimension index, and *d* represents the feature dimension.

To model temporal correlations within each context, we introduce a mutual intra-context attention (MI-Attn) mechanism that enables bidirectional interactions between the context token and its corresponding feature sequence. Specifically, the context token captures temporal dependencies within the feature sequence to build a compact contextual representation, while the feature sequence is further refined through its interaction with the context token and temporal information exchange among sequence elements. As a result, MI-Attn jointly improves context-level modelling and fine-grained temporal representations. Moreover, MI-Attn employs multiple parallel attention heads to capture complementary dependencies from different representation subspaces, thereby learning richer and more discriminative temporal patterns. For the *n*-th attention head, the temporal attention is computed as follows:

$$\alpha_{i,n}^{\text{MI-Attn}} = \text{Softmax}\left(\frac{Q'_{i,n} \cdot K'^{\,T}_{i,n}}{\sqrt{d}}\right) \tag{3}$$

where $Q'_{i,n} = F_i' \cdot W_{i,n}^{Q}$ and $K'_{i,n} = F_i' \cdot W_{i,n}^{K}$ are query and key of context $i$, where $W_{i,n}^{Q}$ and $W_{i,n}^{K}$ are learnable projection matrices. And $\alpha_{i,n}^{\text{MI-Attn}} \in \mathbb{R}^{(N+1)\times(N+1)}$ represents the temporal attention matrix, where each entry represents the attention weight between a pair of sequence elements.

By multiplying $V'_{i,n}$ with $\alpha_{i,n}^{\text{MI-Attn}}$, the context token aggregates temporal information from the entire sequence, while individual frame features are simultaneously enriched:

$$\tilde{F}_{i,n} = \alpha_{i,n}^{\text{MI-Attn}} \cdot V'_{i,n} \tag{4}$$

where $V'_{i,n} = F_i' \cdot W_{i,n}^{V}$ is the value of context $i$, and $W_{i,n}^{V}$ is a learnable projection matrix.

The outputs from all attention heads are concatenated and linearly projected to obtain the final ICF module output $\tilde{F}_i = [\tilde{f}_i^{cls}, \tilde{f}_i^{t-N+1}, \ldots, \tilde{f}_i^{t}]$.

### 4.3 Cross-Context Fusion

The cross-context fusion (CCF) module enables feature interaction across different contexts, facilitating early-stage multi-context fusion and the integration of complementary information from different contexts. Specifically, all updated context tokens from the ICF module are extracted and combined with a learnable global CLS token, which serves as a compact global representation of the multi-context information. To further promote information exchange across contexts and enable global multi-context summarization, the context token sequence $F_C = \left[ f_g^{cls}, \tilde{f}_P^{cls}, \tilde{f}_L^{cls}, \tilde{f}_V^{cls}, \tilde{f}_E^{cls} \right]$ is then fed into the mutual cross-context attention (MC-Attn) mechanism, which is designed to capture inter-context correlations. Moreover, MC-Attn employs multiple attention heads to perform parallel modelling in different representation subspaces. For the $m$-th attention head, the cross-context attention is computed as follows:

$$\alpha_m^{\text{MC-Attn}} = \text{Softmax}\left(\frac{\tilde{Q}_m \cdot \tilde{K}_m^{T}}{\sqrt{d}}\right) \tag{5}$$

where $\tilde{Q}_m$ and $\tilde{K}_m$ are query and key matrices obtained by applying learnable projections $\tilde{W}_m^Q$ and $\tilde{W}_m^K$ to $F_C$. $\alpha_m^{\text{MC-Attn}} \in \mathbb{R}^{5\times 5}$ denotes the cross-context attention matrix, capturing pairwise attention weights among different contexts.

By multiplying $\tilde{V}_m$ with $\alpha_m^{\text{MC-Attn}}$, the global CLS token integrates information from all context tokens, enabling early-stage multi-context fusion, while the individual context tokens simultaneously exchange and incorporate information from other contexts:

$$\tilde{F}'_{C,m} = \alpha_m^{\text{MC-Attn}} \cdot \tilde{V}_m \tag{6}$$

where $\tilde{V}_m$ is the value matric obtained by applying learnable projections $\tilde{W}_m^V$ to $F_C$.

The outputs from all attention heads are then concatenated and linearly projected to obtain the final CCF module output $\tilde{F}'_C = \left[\tilde{f}'^{cls}_{g}, \tilde{f}'^{cls}_{P}, \tilde{f}'^{cls}_{L}, \tilde{f}'^{cls}_{V}, \tilde{f}'^{cls}_{E}\right]$.

### 4.4 Intra-Context Refinement

The intra-context refinement (ICR) module performs guided refinement of individual context tokens, which summarize context-specific information over the entire temporal sequence, thereby enabling more effective final-stage feature integration within each context. For context $i$, the updated context token $\tilde{f}'^{cls}_i$ from the CCF module retain their inherent contextual information while simultaneously integrating cross-contextual knowledge, which is further propagated during intra-context refinement. $\tilde{f}'^{cls}_i$ is then prepended to the contextual feature sequences output from the ICF module. These sequences are then processed by a feed-forward network (FFN) with residual connections and normalization, yielding refined contextual sequences $\bar{F}_i = [\bar{f}_i^{cls}, \bar{f}_i^{t-N+1}, \ldots, \bar{f}_i^{t}]$.

The core of the ICR module is a guided intra-context attention (GI-Attn) mechanism, which aggregates temporal information within each context to further refine the corresponding context token. Notably, unlike MI-Attn, which performs bidirectional interactions between the context token and the feature sequence, GI-Attn operates in a directed manner, where only the context token attends to the feature sequence for refinement. GI-Attn consists of multiple attention heads. For $h$-th attention head, the query is generated by linearly projecting the context token $\bar{f}_i^{cls}$, whereas the key and value are obtained from the contextual feature sequence:

$$\bar{Q}_h = \bar{f}_i^{cls} \cdot \bar{W}^{Q}_{i,h}, \quad \bar{K}_h = \bar{F}_i \cdot \bar{W}^{K}_{i,h}, \quad \bar{V}_h = \bar{F}_i \cdot \bar{W}^{V}_{i,h} \tag{7}$$

where $\bar{Q}_h$, $\bar{K}_h$, and $\bar{V}_h$ denote the query, key, and value representations of the $h$-th attention head, respectively, and $\bar{W}^{Q}_{i,h}$, $\bar{W}^{K}_{i,h}$, and $\bar{W}^{V}_{i,h}$ are learnable projection matrices.

The guided intra-context attention map is computed as:

$$\alpha_h^{\text{GI-Attn}} = \text{Softmax}\left(\frac{\bar{Q}_h \cdot \bar{K}_h^{T}}{\sqrt{d}}\right) \tag{8}$$

where $\alpha_h^{\text{GI-Attn}} \in \mathbb{R}^{1\times(N+1)}$ denotes the guided intra-context attention matrix, representing the attention weights assigned by the context token to individual elements of the corresponding feature sequence.

The context token is further refined by multiplying $\alpha_h^{\text{GI-Attn}}$ with $\bar{V}_h$:

$$\hat{f}^{cls}_{i,h} = \alpha_h^{\text{GI-Attn}} \cdot \bar{V}_h \tag{9}$$

The refined context token $\hat{f}^{cls}_i$ is obtained by concatenating the outputs of all attention heads, followed by a linear projection.

### 4.5 Cross-Context Refinement

The cross-context refinement (CCR) module serves as the final-stage multi-context fusion module, consolidating directed cross-context interactions and further refining the global CLS token, which encodes the final global representation. The core of the module is the guided cross-context attention (GC-Attn) mechanism, through which the global CLS token attends to the context tokens to integrate complementary information across contexts. Specifically, the global CLS token $\tilde{f}'^{cls}_g$ generated by the CCF module is concatenated with the updated context tokens $\hat{f}^{cls}_P, \hat{f}^{cls}_L, \hat{f}^{cls}_V, \hat{f}^{cls}_E$ from the ICR module to form context sequence $\hat{F}_C$, which is passed into the GC-Attn. Notably, unlike MC-Attn, which enables bidirectional information exchange between the global CLS token and the context tokens, GC-Attn adopts a directed cross-context fusion scheme in which information flows only from the context sequence $\hat{F}_C$ to the global CLS token. The architecture of GC-Attn is shown in Fig. 3. GC-Attn consists of multiple attention heads. For the $l$-th attention head, the global CLS token is used to generate the query, while the key and value

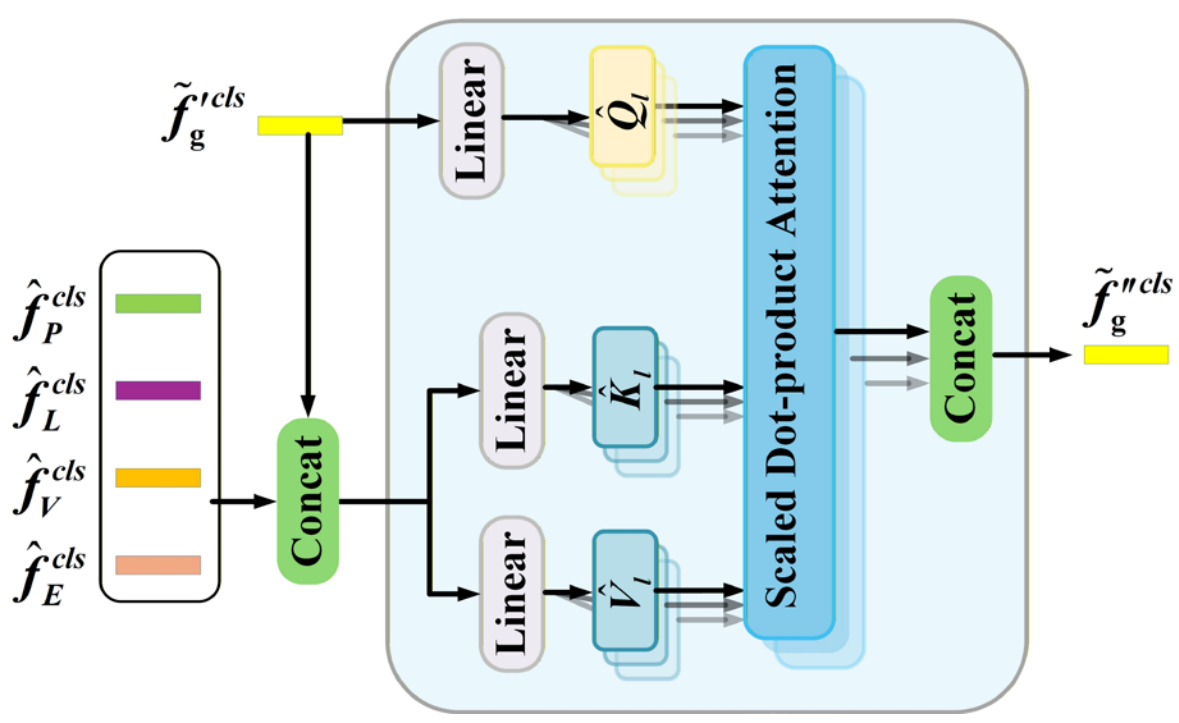

**Fig. 3** The architecture of the guided cross-context attention

are derived from $\hat{F}_C$:

$$\hat{Q}_l = \tilde{f}'^{cls}_{g} \cdot \hat{W}_l^{Q}, \quad \hat{K}_l = \hat{F}_C \cdot \hat{W}_l^{K}, \quad \hat{V}_l = \hat{F}_C \cdot \hat{W}_l^{V} \tag{10}$$

where $\hat{Q}_l$, $\hat{K}_l$, and $\hat{V}_l$ denote the query, key, and value representations of the $l$-th attention head, respectively, and $\hat{W}_l^{Q}$, $\hat{W}_l^{K}$, and $\hat{W}_l^{V}$ are learnable projection matrices.

The guided cross-context attention is computed as:

$$\alpha_l^{\text{GC-Attn}} = \text{Softmax}\left(\frac{\hat{Q}_l \cdot \hat{K}_l^{T}}{\sqrt{d}}\right) \tag{11}$$

where $\alpha_l^{\text{GC-Attn}} \in \mathbb{R}^{1\times 5}$ denotes the guided cross-context attention matrix, which encodes the attention weights assigned by the global CLS token to individual elements in context sequence $\hat{F}_C$.

Finally, the global CLS token is further refined by multiplying $\alpha_l^{\text{GC-Attn}}$ with $\hat{V}_l$, thereby performing selective cross-context aggregation to form a unified global representation:

$$\tilde{f}''^{cls}_{g,l} = \alpha_l^{\text{GC-Attn}} \cdot \hat{V}_l \tag{12}$$

The refined global CLS token $\tilde{f}''^{cls}_{g}$, formed via concatenation of the outputs from all attention heads and a subsequent linear projection, is subsequently processed by a multi-layer perceptron (MLP) to produce the pedestrian intention.

## 5. Experiments

### 5.1 Datasets and Implementation Details

MFT is evaluated on two public benchmarks, JAAD [2] and PIE [3]. JAAD dataset contains 346 HD videos from North America and Eastern Europe. We use its two annotation variants, $JAAD_{beh}$ and $JAAD_{all}$ (the latter includes 2,100 additional non-crossing pedestrians). Following [5], the dataset is split into 177, 117, and 29 videos for training, testing, and validation. From the annotated sequences, 16-frame clips are sampled with 80% overlap. PIE dataset consists of 6 hours of HD video recorded in Toronto with frame-level annotations for 1,842 pedestrians. Following [5], we use set01, set02, and set06 for training, set04 and set05 for validation, and set03 for testing. For the PIE dataset, the overlap ratio is set to 60%.

The number of attention heads is set to 4 for all multi-head attention mechanisms. Both the embedding space and the Transformer hidden dimension are configured to 128. A dropout rate of 0.2 is applied in the MLP. The batch size is set to 2, and the model is trained for a total of 60 epochs. We employ binary cross-entropy loss and optimize the model using Adam with a learning rate of $5\times10^{-7}$ for JAAD dataset and $2\times10^{-5}$ for PIE dataset. To mitigate class imbalance, we incorporate class weights proportional to the positive–negative sample ratio.

### 5.2 Quantitative Results

We compare MFT with several state-of-the-art (SOTA) methods. The results on JAAD and PIE datasets are shown in Table 1. Green, blue and orange indicate the best, second-best and third-best results, respectively. The final row ("Improvement") summarizes the relative performance gain or degradation of MFT compared with the SOTA methods. We evaluate performance with five commonly used metrics: accuracy (Acc), area under the curve (AUC), F1 score, precision (P), and recall (R), consistent with the benchmark settings [5]. For MFT, all metrics are reported as mean ± standard deviation over five independent runs with different random seeds.

On the $JAAD_{beh}$ dataset, MFT achieves the highest accuracy of 0.738 ± 0.011, outperforming the second-best meth-

**Table 1.** Quantitative results on JAAD and PIE dataset

| Models | $JAAD_{beh}$ | | | | | $JAAD_{all}$ | | | | | PIE | | | | |
|---|---|---|---|---|---|---|---|---|---|---|---|---|---|---|---|
| | Acc | AUC | F1 | P | R | Acc | AUC | F1 | P | R | Acc | AUC | F1 | P | R |
| IntFormer [15] | 0.59 | 0.54 | 0.69 | - | - | 0.86 | 0.78 | 0.62 | - | - | 0.89 | 0.92 | 0.81 | - | - |
| PCPA [5] | 0.56 | 0.54 | 0.63 | 0.66 | 0.60 | 0.77 | 0.79 | 0.56 | 0.42 | 0.83 | 0.87 | 0.85 | 0.78 | 0.76 | 0.81 |
| Global PCPA [4] | 0.62 | 0.54 | 0.74 | 0.65 | 0.85 | 0.83 | 0.82 | 0.63 | 0.51 | 0.81 | 0.89 | 0.86 | 0.80 | 0.79 | 0.81 |
| MMH-PAP [16] | - | - | - | - | - | 0.84 | 0.80 | 0.62 | 0.54 | - | 0.89 | 0.88 | 0.81 | 0.77 | - |
| TrouSPI-Net [17] | 0.64 | 0.56 | 0.76 | 0.66 | 0.91 | 0.85 | 0.73 | 0.56 | 0.57 | 0.55 | 0.88 | 0.88 | 0.80 | 0.73 | 0.89 |
| STCrossingPose [18] | 0.63 | 0.56 | 0.74 | 0.66 | 0.83 | - | - | - | - | - | - | - | - | - | - |
| V-PedCross [10] | 0.61 | 0.50 | 0.75 | 0.71 | 0.80 | 0.82 | 0.74 | 0.64 | 0.58 | 0.63 | 0.89 | 0.88 | 0.67 | 0.74 | 0.84 |
| RU-LSTM [20] | 0.69 | 0.62 | 0.78 | - | - | 0.86 | 0.78 | 0.62 | - | - | 0.87 | 0.84 | 0.77 | - | - |
| STFF-MANet [21] | 0.66 | 0.58 | 0.77 | 0.67 | 0.89 | 0.89 | 0.80 | 0.67 | 0.68 | 0.67 | 0.89 | 0.88 | 0.82 | 0.79 | 0.85 |
| MTMGN [22] | 0.70 | 0.70 | 0.83 | 0.79 | 0.87 | 0.89 | 0.89 | 0.73 | 0.66 | 0.89 | 0.90 | 0.87 | 0.92 | 0.95 | 0.90 |
| Dual-STGAT [23] | - | - | - | - | - | 0.92 | 0.92 | 0.73 | 0.86 | 0.63 | 0.86 | 0.87 | 0.91 | 0.92 | 0.90 |
| MTC [32] | 0.71 | 0.65 | 0.80 | 0.72 | 0.90 | 0.90 | 0.82 | 0.70 | 0.70 | 0.70 | - | - | - | - | - |
| LSOP-Net [24] | 0.65 | 0.54 | 0.78 | 0.65 | 0.98 | 0.85 | 0.75 | 0.58 | 0.56 | 0.61 | 0.89 | 0.87 | 0.81 | 0.80 | 0.82 |
| ACIT [33] | 0.70 | 0.62 | 0.79 | 0.69 | 0.93 | 0.89 | 0.83 | 0.69 | 0.65 | 0.74 | - | - | - | - | - |
| PFRN [19] | - | - | - | - | - | - | - | - | - | - | 0.90 | 0.85 | 0.77 | 0.81 | 0.74 |
| GTransPDM [34] | - | - | - | - | - | 0.87 | 0.78 | 0.64 | 0.64 | | 0.90 | 0.87 | 0.82 | 0.86 | |
| SAM [12] | 0.67 | 0.60 | 0.77 | 0.68 | 0.90 | 0.83 | 0.80 | 0.62 | 0.47 | 0.82 | - | - | - | - | - |
| RAIDN [36] | 0.70 | 0.68 | 0.75 | 0.75 | 0.77 | 0.89 | 0.80 | 0.66 | 0.65 | 0.72 | 0.92 | 0.89 | 0.85 | 0.82 | 0.89 |
| Faster-PCPNet [37] | - | - | - | - | - | 0.89 | 0.77 | 0.65 | 0.73 | 0.58 | 0.94 | 0.92 | 0.89 | 0.89 | 0.88 |
| MFT | 0.738 ±0.011 | 0.701 ±0.008 | 0.803 ±0.009 | 0.756 ±0.006 | 0.857 ±0.016 | 0.933 ±0.013 | 0.955 ±0.007 | 0.838 ±0.027 | 0.728 ±0.043 | 0.990 ±0.002 | 0.899 ±0.005 | 0.885 ±0.031 | 0.819 ±0.008 | 0.828 ±0.025 | 0.812 ±0.025 |
| Improvement (↑↓) | ↑0.03 | ↑0.02 | ↓0.03 | ↓0.03 | ↓0.12 | ↑0.01 | ↑0.04 | ↑0.11 | ↓0.13 | ↑0.10 | ↓0.04 | ↓0.03 | ↓0.10 | ↓0.12 | ↓0.09 |

Note: The results of the SOTA methods are taken from their original publications. A dash (–) indicates that the corresponding metric was not reported.

od, MTC [32], by approximately 0.03 and surpassing MTMGN [22], ACIT [33], and RAIDN [36] by approximately 0.04. MFT also achieves the highest AUC of 0.701 ± 0.008, which is comparable to MTMGN [22] and exceeds the second-best method, RAIDN [36], by 0.02. In terms of F1 score and precision, MFT ranks second, achieving 0.803 ± 0.009 and 0.756 ± 0.006, respectively. MFT achieves an F1 score comparable to that of MTC [32] but approximately 0.03 lower than MTMGN [22], while its precision is 0.03 below MTMGN [22] and 0.01 above RAIDN [36]. The recall performance is relatively modest, reaching 0.857 ± 0.016, approximately 0.12 below the best recall of 0.98 reported by LSOP-Net [24].

On the JAAD$_{all}$ dataset, MFT achieves competitive performance, ranking first in accuracy, AUC, F1 score, and recall, with values of 0.933 ± 0.013, 0.955 ± 0.007, 0.838 ± 0.027, 0.990 ± 0.002, respectively. In terms of accuracy, MFT surpasses Dual-STGAT [23] by approximately 0.01 and MTC [32] by approximately 0.03. MFT also achieves substantial improvements in AUC and F1 score, outperforming Dual-STGAT [23] by approximately 0.04 in AUC and surpassing MTMGN [22] and Dual-STGAT [23] by approximately 0.11 in F1 score. Its recall exceeds the second-best method MTMGN [22] by approximately 0.10. MFT ranks third in precision, falling 0.13 behind the top-performing method, Dual-STGAT [23].

On the PIE dataset, MFT remains competitive, achieving accuracy of 0.899 ± 0.005 and ranking third, behind Faster-PCPNet [37] and RAIDN [36] by approximately 0.04 and 0.02, respectively. MFT achieves an AUC of 0.885 ± 0.031, which is comparable to the 0.89 reported by RAIDN [36] and approximately 0.03 lower than the best AUC of 0.92 reported by IntFormer [15] and Faster-PCPNet [37].

It should be noted that MFT and the compared methods generally rely on the same underlying data sources, namely RGB images and vehicle speed, but use different input representations derived through different preprocessing pipelines. Existing methods may directly process RGB images or use derived input modalities, such as pose sequences, optical flow, depth maps, and semantic maps, which are generated through additional preprocessing or pretrained models. In contrast, MFT employs compact high-level semantic attributes as input representations. Therefore, the comparison evaluates the overall prediction performance of methods using different representations derived from comparable underlying data sources. MFT is evaluated using the high-level attribute annotations provided by the benchmark datasets. In practical deployment, these attributes can be generated online by vehicle-side or roadside perception modules, enabling MFT to operate within vehicle–infrastructure cooperative intelligent transportation systems, such as the one presented in [44].

Despite these differences in input representation, the quantitative results demonstrate that MFT effectively integrates compact, explicitly defined semantic attributes from pedestrian behavior, localization, vehicle motion, and the traffic environment, thereby capturing their complementary contextual cues. Its progressive fusion strategy facilitates the multi-stage integration of intra- and cross-context information through dedicated attention mechanisms, improving the effectiveness of contextual feature extraction and fusion. Furthermore, the compact global CLS token captures global semantic and contextual cues, contributing to MFT's competitive or superior performance relative to SOTA methods across the evaluated benchmarks.

### 5.3 Ablation Studies

We conduct an ablation study to investigate the effects of different context combinations on MFT's prediction performance. The base model incorporates four types of context, namely pedestrian behavior context (*P*), pedestrian localization context (*L*), vehicle motion context (*V*), and environmental context (*E*). To analyze the effect of different contexts and their combinations, we progressively remove one, two, or three types of contextual inputs from the base model. Among them, pedestrian localization context is always retained, as it is essential for identifying the target pedestrian in the scene. The results are shown in Table 2. Green, blue and orange indicate the best, second-best and third-best results, respectively.

**Table 2.** Ablation study of different context configurations

| Context Configuration | | | | JAAD$_{beh}$ | | | | | JAAD$_{all}$ | | | | | PIE | | | | |
|---|---|---|---|---|---|---|---|---|---|---|---|---|---|---|---|---|---|---|
| L | P | V | E | Acc | AUC | F1 | P | R | Acc | AUC | F1 | P | R | Acc | AUC | F1 | P | R |
| ✓ | ✗ | ✓ | ✓ | 0.68 | 0.69 | 0.78 | 0.70 | 0.87 | 0.90 | 0.96 | 0.78 | 0.65 | 0.96 | 0.88 | 0.93 | 0.80 | 0.74 | 0.87 |
| ✓ | ✓ | ✗ | ✓ | 0.70 | 0.67 | 0.77 | 0.75 | 0.79 | 0.91 | 0.94 | 0.79 | 0.65 | 0.99 | 0.83 | 0.80 | 0.71 | 0.68 | 0.74 |
| ✓ | ✓ | ✓ | ✗ | 0.69 | 0.77 | 0.68 | 0.93 | 0.53 | 0.88 | 0.94 | 0.75 | 0.60 | 1.00 | 0.87 | 0.92 | 0.79 | 0.73 | 0.87 |
| ✓ | ✗ | ✗ | ✓ | 0.61 | 0.49 | 0.76 | 0.62 | 0.97 | 0.89 | 0.93 | 0.77 | 0.62 | 0.99 | 0.77 | 0.70 | 0.57 | 0.60 | 0.55 |
| ✓ | ✓ | ✗ | ✗ | 0.66 | 0.61 | 0.75 | 0.69 | 0.83 | 0.90 | 0.90 | 0.76 | 0.66 | 0.89 | 0.81 | 0.79 | 0.69 | 0.64 | 0.75 |
| ✓ | ✗ | ✓ | ✗ | 0.63 | 0.47 | 0.76 | 0.64 | 0.93 | 0.76 | 0.84 | 0.53 | 0.41 | 0.76 | 0.83 | 0.91 | 0.75 | 0.65 | 0.87 |
| ✓ | ✗ | ✗ | ✗ | 0.60 | 0.48 | 0.75 | 0.62 | 0.96 | 0.72 | 0.85 | 0.52 | 0.37 | 0.86 | 0.75 | 0.71 | 0.59 | 0.64 | 0.64 |
| ✓ | ✓ | ✓ | ✓ | 0.74 | 0.70 | 0.80 | 0.76 | 0.86 | 0.93 | 0.96 | 0.84 | 0.73 | 0.99 | 0.90 | 0.89 | 0.82 | 0.83 | 0.81 |

**Table 3.** Ablation study of different attention mechanisms in the cross-context refinement module

| Variants | Fusion Mechanism | JAAD$_{beh}$ | | | | | JAAD$_{all}$ | | | | | PIE | | | | |
|---|---|---|---|---|---|---|---|---|---|---|---|---|---|---|---|---|
| | | Acc | AUC | F1 | P | R | Acc | AUC | F1 | P | R | Acc | AUC | F1 | P | R |
| MFT-V1 | Average Pooling | 0.70 | 0.73 | 0.79 | 0.72 | 0.87 | 0.91 | 0.98 | 0.79 | 0.66 | 0.99 | 0.89 | 0.94 | 0.81 | 0.83 | 0.78 |
| MFT-V2 | Additive Attention [4] | 0.72 | 0.70 | 0.79 | 0.76 | 0.82 | 0.92 | 0.96 | 0.81 | 0.69 | 0.99 | 0.89 | 0.93 | 0.81 | 0.81 | 0.81 |
| MFT-V3 | Gated Attention [36] | 0.71 | 0.66 | 0.79 | 0.72 | 0.88 | 0.90 | 0.93 | 0.78 | 0.65 | 0.97 | 0.90 | 0.88 | 0.82 | 0.80 | 0.85 |
| MFT-V4 | Inter-Modal Attention [32] | 0.72 | 0.67 | 0.79 | 0.73 | 0.87 | 0.91 | 0.94 | 0.79 | 0.65 | 1.00 | 0.88 | 0.86 | 0.79 | 0.77 | 0.81 |
| MFT | Guided Cross-Context Attention | 0.74 | 0.70 | 0.80 | 0.76 | 0.86 | 0.93 | 0.96 | 0.84 | 0.73 | 0.99 | 0.90 | 0.89 | 0.82 | 0.83 | 0.81 |

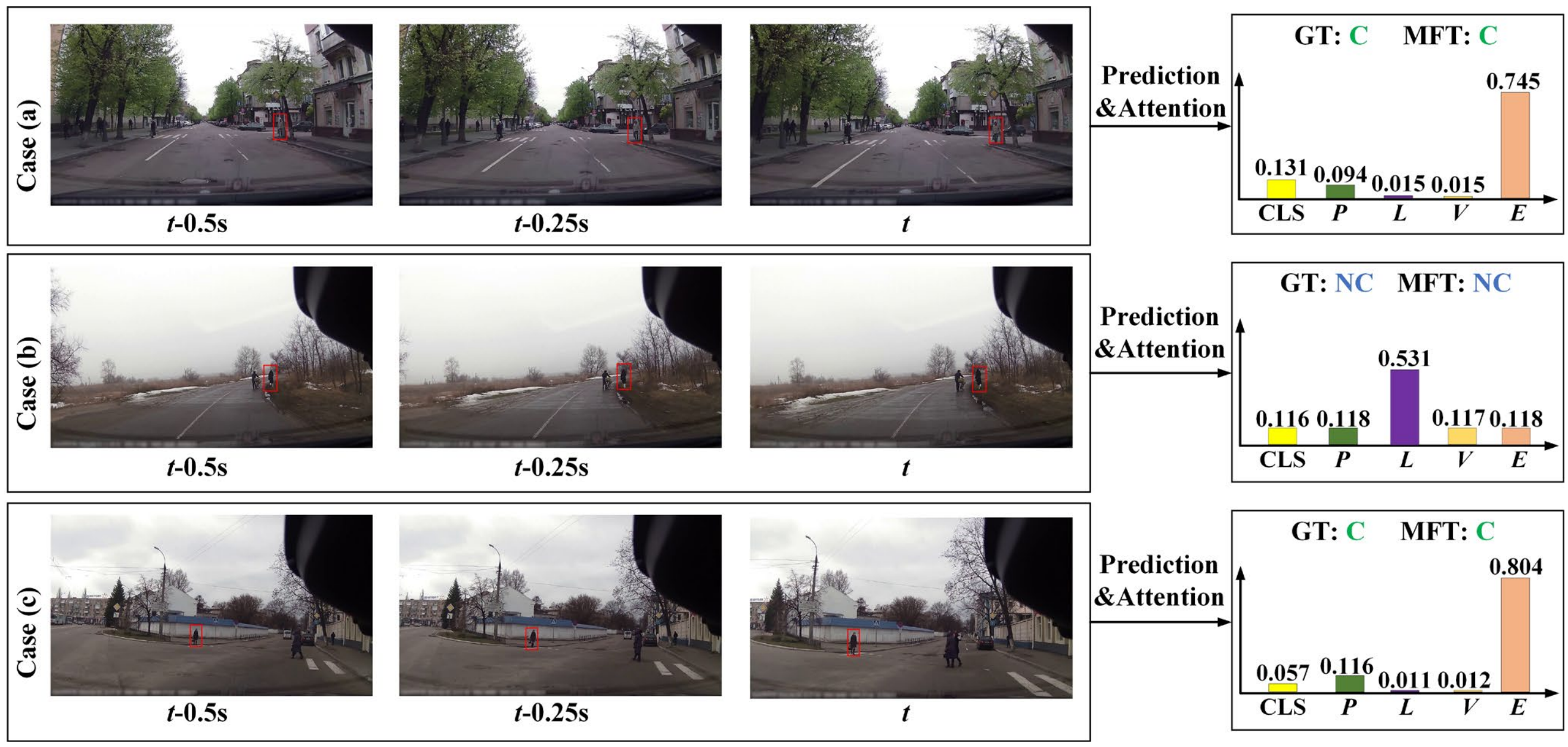

**Fig. 4** Qualitative results of MFT network

The results show that removing a single type of contextual input causes only a modest performance degradation, with drops of 0.04 – 0.06 on JAAD$_{beh}$ dataset, 0.02 – 0.05 on JAAD$_{all}$ dataset, and 0.02 – 0.07 on PIE dataset. In contrast, removing two types of contextual inputs leads to a much more severe deterioration in performance. For example, when only $L$ and $E$ are used, the accuracy on JAAD$_{beh}$ dataset drops by 0.13, highlighting the importance of $P$ and $V$. On JAAD$_{all}$ dataset, using only $L$ and $V$ results in a 0.17 decrease in accuracy, while on PIE dataset, using only $L$ and $E$ causes a 0.13 drop. When relying solely on $L$, the performance drops to its lowest level, decreasing by 0.14, 0.21, and 0.15 compared with the base model on JAAD$_{beh}$, JAAD$_{all}$, and PIE datasets, respectively. Overall, the results highlight that, in addition to pedestrian localization, pedestrian behavior, vehicle motion, and environmental contexts provide complementary semantic and dynamic information that cannot be captured by pedestrian localization alone, and their effective integration is essential for accurately modelling pedestrian intention.

We also design several variants of our network to investigate the effectiveness of different fusion mechanisms in the CCR module, as this module plays a crucial role in fusing information from multiple contexts into the final global representation. Specifically, we replace the proposed GC-Attn with four alternative fusion mechanisms: (1) MFT-V1: all context tokens in the CCR module are fused by average pooling. (2) MFT-V2: the context tokens are fused using the additive attention mechanism adopted in [4], where the attention weights are computed via a learnable additive compatibility function. (3) MFT-V3: the context tokens are fused using a gated attention mechanism similar to that in [36], where global pooling is first applied to generate modulation weights, followed by attention-based adaptive fusion. (4) MFT-V4: the context tokens are refined by attending to one another, following the inter-modal attention mechanism in [32], and the refined tokens are then concatenated for fusion. The results are shown in Table 3, where the best results are highlighted in green.

It can be observed that MFT-V1 results in a performance drop, with accuracy decreasing by 0.04, 0.02, and 0.01, on JAAD$_{beh}$, JAAD$_{all}$, and PIE datasets, respectively. This may be attributed to the use of average pooling, which treats all context tokens equally and fails to capture the fine-grained contributions of each context token as effectively as the proposed GC-Attn. MFT-V2 also leads to a decline in accuracy, suggesting that additive attention is insufficient for modelling complex relationships among multiple context tokens. Its scalar weighting and linear aggregation limit the modelling of fine-grained contributions and inter-context dependencies compared with the proposed GC-Attn. MFT-V3 leads to lower accuracy on JAAD$_{beh}$ and JAAD$_{all}$ datasets, while achieving comparable performance on PIE dataset. This may be because gated attention compresses interaction features through global average pooling and performs channel-wise reweighting, which can cause information loss and fails to explicitly model inter-context relationships. MFT-V4 shows inferior performance because its unguided interactions among context tokens may introduce redundant information, making it less effective than the proposed GC-Attn for modelling relevant cross-context dependencies.

### 5.4 Qualitative Results

We present qualitative results for several illustrative cases, where the target pedestrian is enclosed within a red rectangular boundary. Ground truth is indicated by GT, with C and NC representing crossing and non-crossing, respectively. The results are shown in Fig. 4, where the attention scores of GC-Attn in the CCR module, averaged across all attention heads, are visualized to illustrate how the global CLS token attends to different context tokens in the corresponding scenes.

The results show that MFT can reliably predict pedestrian's intentions across the following cases: In case (a), the pedestrian remains stationary at the roadside in front of a crosswalk, exhibiting no body movement but maintaining eye contact with the vehicle. Under these conditions, MFT successfully predicts the intention to cross, with attention weights primarily focused on the environmental context $E$ (74.5%) and the global CLS token capturing early-stage holistic information (13.1%), while a non-negligible portion is allocated to the pedestrian behavior context $P$ (9.4%). This outcome suggests that MFT effectively integrates environmental cues such as crosswalk availability with subtle behavioral indicators, thereby enabling robust intention prediction even in the absence of immediate motion cues. In case (b), the pedestrian is walking longitudinally along the roadside. MFT accurately predicts a non-crossing intention, with attention primarily focused on the pedestrian localization context $L$ (53.1%), which reflects the pedestrian's consistent lateral position along the road, while the other contexts receive similar levels of attention. In case (c), the scene depicts a woman walking from the roadside toward the crosswalk, accompanied by a pedestrian sign. MFT accurately infers her crossing intention by focusing predominantly on cues from the environmental context $E$ (80.4%), thereby emphasizing the pivotal contribution of traffic infrastructure cues to intention prediction.

### *5.5 Computational Cost Analysis*

We evaluate the real-time performance and computational cost of MFT by comparing it with several representative methods, including PCPA [5], Global PCPA [4], FUSSI [28], MFFN [29], MTC [32], VMI [26], ACIT [33], and MTMGN [22]. For evaluation, we test the model trained on the JAAD dataset as an example. The results are presented in Table 4, where green and blue highlight the best and second-best performance, respectively. MFT achieves the smallest parameter number of 0.95 million, lower than the second-smallest model, FUSSI [28] (1.00 million). Regarding model size, MFT ranks second at 9.40 MB, slightly larger than FUSSI [28] (8.40 MB) but still smaller than all other methods. In terms of inference time, MFT attains the second-best result of 23.20 ms, slower than VMI [26] (11.03 ms) but notably faster than FUSSI [28] (34.92 ms), which ranks third. The results of the computational cost analysis demonstrate that MFT exhibits competitive model compactness and real-time inference performance.

**Table 4.** Comparison of computational cost

| Models | Size (MB) | Parameters (Million) | Inference Time (ms) |
| --- | --- | --- | --- |
| PCPA [5] | 118.80 | 31.17 | 38.60 |
| Global PCPA [4] | 374.20 | 60.92 | 70.83 |
| FUSSI [28] | 8.40 | 1.00 | 34.92 |
| MFFN [29] | - | - | 46.20 |
| MTC [32] | 99.70 | 8.25 | 36.23 |
| VMI [26] | 19.07 | - | 11.03 |
| ACIT [33] | 62.50 | 5.15 | 43.93 |
| MTMGN [22] | - | - | 56.00 |
| MFT | 9.40 | 0.95 | 23.20 |

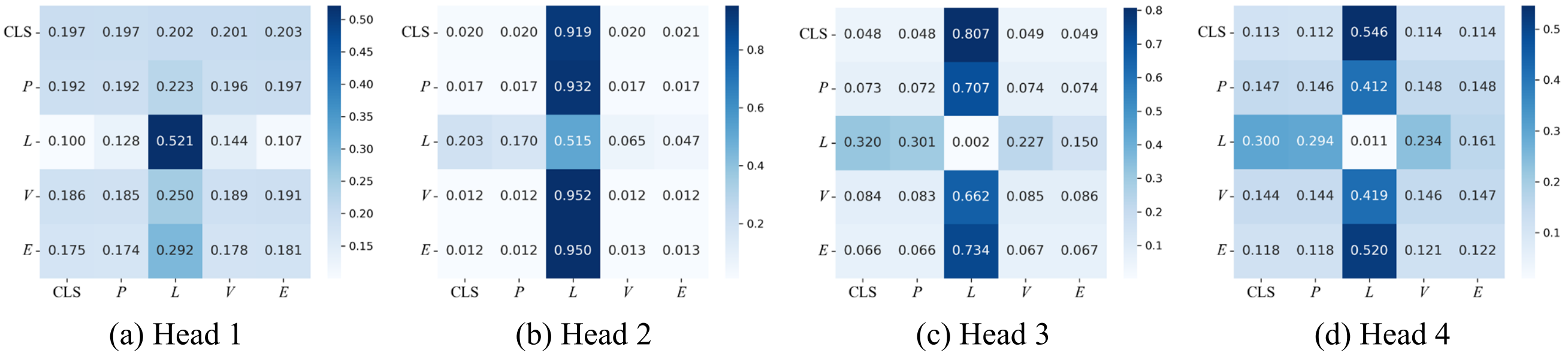

(a) Head 1 (b) Head 2 (c) Head 3 (d) Head 4

**Fig. 5** Visualization of attention maps generated by the MC-Attn mechanism on the $\text{JAAD}_{\text{beh}}$ dataset

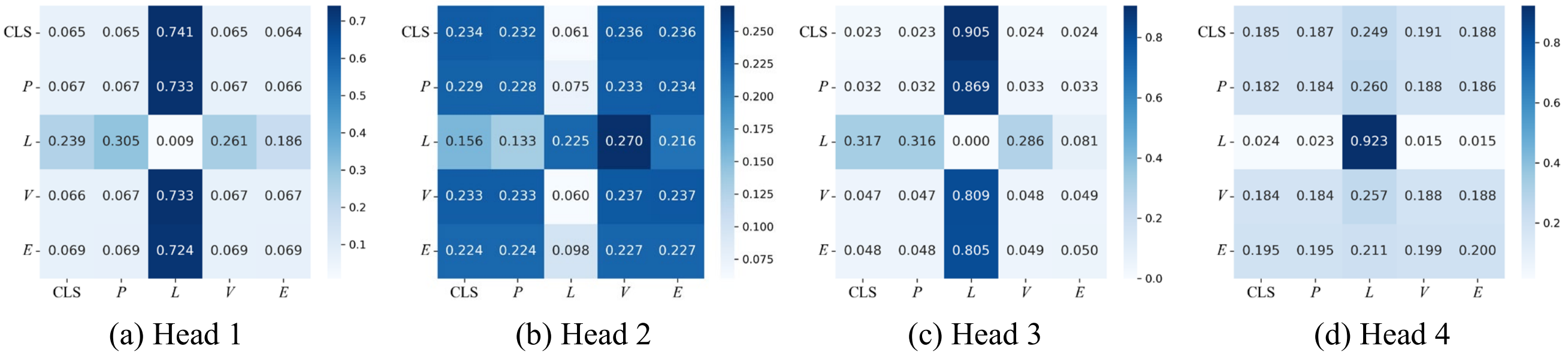

(a) Head 1 (b) Head 2 (c) Head 3 (d) Head 4

**Fig. 6** Visualization of attention maps generated by the MC-Attn mechanism on the $\text{JAAD}_{\text{all}}$ dataset

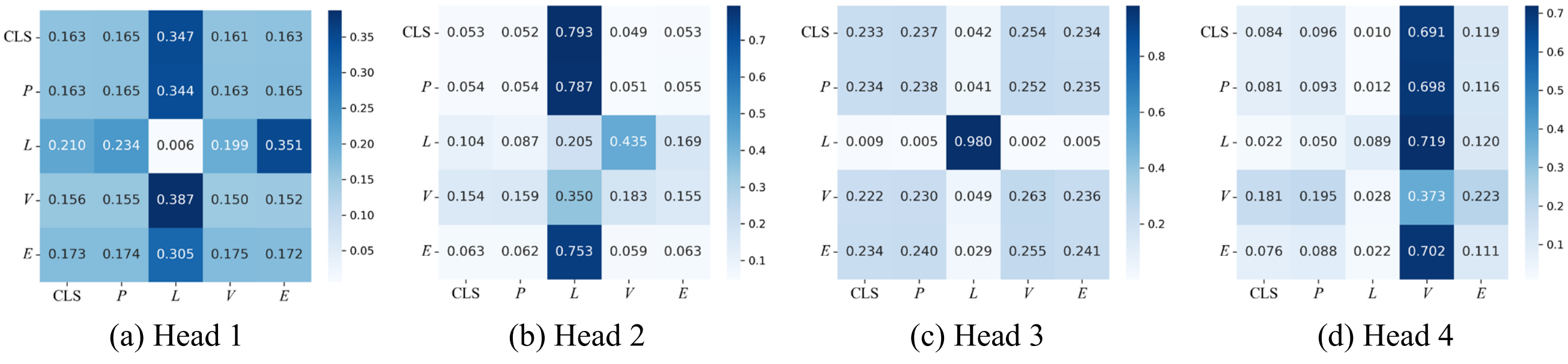

(a) Head 1 (b) Head 2 (c) Head 3 (d) Head 4

**Fig. 7** Visualization of attention maps generated by the MC-Attn mechanism on the PIE dataset

### 5.6 Attention Map Visualization

To assess the importance of different contexts at different fusion stages, we visualize the multi-head attention maps of the MC-Attn mechanism in the CCF module and the GC-Attn mechanism in the CCR module across all test samples.

Fig. 5-7 illustrate the multi-head attention maps of MC-Attn mechanism on $\text{JAAD}_{beh}$, $\text{JAAD}_{all}$, and PIE datasets, respectively. The mutual attention between the global CLS token and the context tokens is clearly depicted, where darker blue indicates stronger attention. In the early stage of fusion, the pedestrian localization context ($L$), representing the bounding box coordinates of the pedestrian, receives the greatest attention from the global CLS token. Specifically, on the $\text{JAAD}_{beh}$ and PIE datasets, three heads exhibit the strongest attention with weights of 91.9%, 80.7%, and 54.6% and 34.7%, 79.3%, and 69.1%, respectively, while on the $\text{JAAD}_{all}$ dataset, two heads dominate with weights of 74.1% and 90.5%. In contrast, the attention of other context tokens is more dispersed, with a greater focus on one another while exhibiting weaker self-attention, thereby complementing the information provided by their own context. These observations reveal two key characteristics at the early stage of fusion: (1) The pedestrian localization context $L$ consistently receives dominant attention from the global CLS token, underscoring its crucial role in grounding intention prediction. (2) Other context tokens interact more with each other than with themselves, indicating complementary information exchange rather than reliance on self-attention.

Fig. 8–10 present the multi-head attention maps of the GC-Attn mechanism, illustrating the attention of the global context on all context tokens in the final stage of fusion. It can be observed that the distribution of attention varies across different datasets, and that each head attends to different aspects. On the $\text{JAAD}_{beh}$ dataset, three heads primarily attend to the environmental context $E$ with weights of 38.7%, 56.5%, and 41.5%, while one head focuses on the global CLS token, capturing early-stage holistic information, with a weight of 42.1%. On the $\text{JAAD}_{all}$ dataset, a comparable pattern emerges: three heads concentrate on the environmental context $E$ with weights of 64.6%, 42.8%, and 40.7%, whereas one head places strong emphasis on the global CLS token with a weight of 84.0%, while the remaining tokens receive a more evenly distributed attention. On the PIE dataset, a different distribution pattern is observed. Only two heads assign attention greater than 40% to specific context tokens, with head 1 attending to the pedestrian behavior context $P$ at 47.0% and head 3 focusing on the vehicle motion context $V$ at 49.3%. The remaining heads exhibit more dispersed attention without a clear domi-

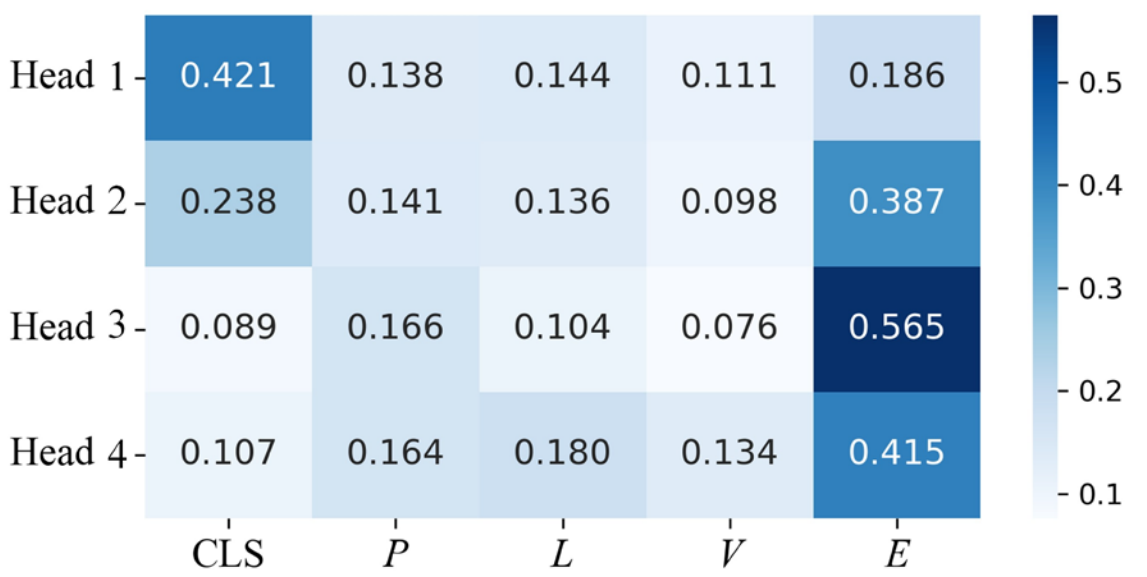

**Fig. 8** Visualization of attention maps generated by the GC-Attn mechanism on the $\text{JAAD}_{beh}$ dataset

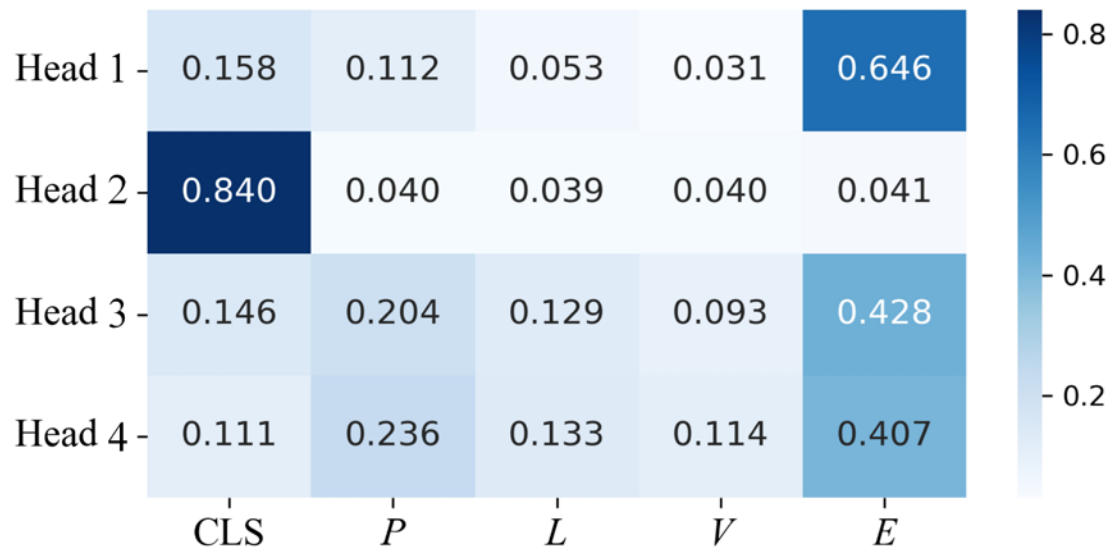

**Fig. 9** Visualization of attention maps generated by the GC-Attn mechanism on the $\text{JAAD}_{all}$ dataset

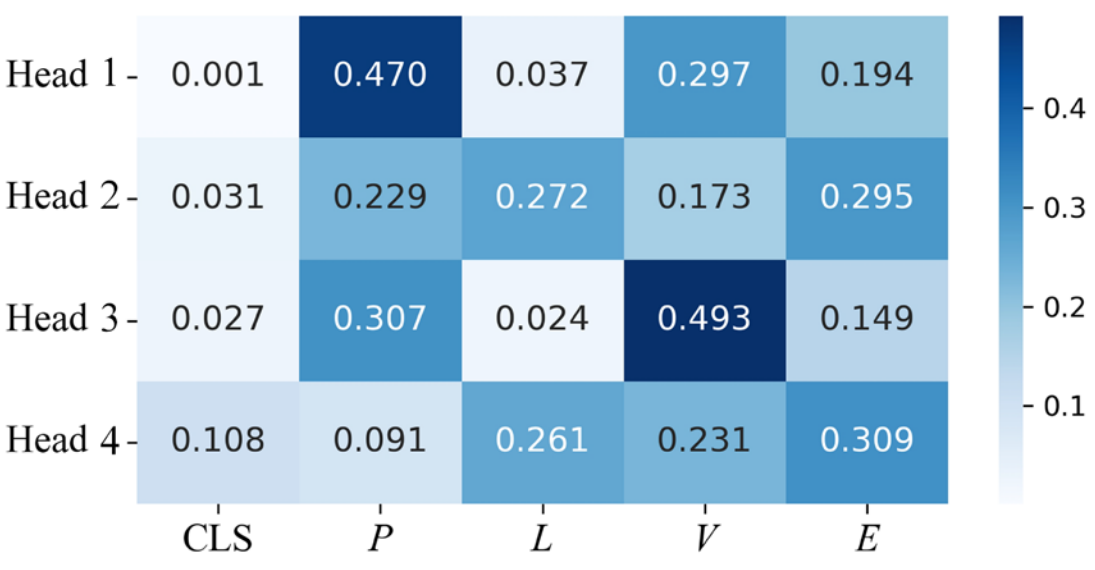

**Fig. 10** Visualization of attention maps generated by the GC-Attn mechanism on the PIE dataset

nant focus. Through the visualization of attention maps, the final-stage fusion reveals not only the varying importance of different contexts across datasets but also the distinct roles played by individual heads. These findings demonstrate that the GC-Attn mechanism effectively captures diverse contextual cues, enabling a more comprehensive understanding of pedestrian intention in complex urban environments.

### 5.7 Effect of Longer Prediction Horizon

While the benchmark setting considers only short prediction horizons of 1–2 seconds [5], we further validate the robustness and effectiveness of the proposed MFT network under extended prediction horizons, thereby assessing its applicability in more challenging scenarios. The time-to-event (TTE) is extended to 2–3 seconds and the resultes on $\text{JAAD}_{beh}$, $\text{JAAD}_{all}$, and PIE datasets are presented in Table 5-7, respectively. It can be observed that a longer TTE leads to decreased model performance, with accuracy dropping from 0.74 to 0.68 on $\text{JAAD}_{beh}$ dataset, from 0.93 to 0.88 on $\text{JAAD}_{all}$ dataset, and from 0.90 to 0.85 on PIE dataset. This decline arises from the inherent difficulty of making reliable long-horizon predictions, given the unpredictability of urban environments and human behavior.

We also compared MFT with Global PCPA [4] and LSOP-Net [23], two raw-modality-based models that exploit implicit cues. The results show that our approach achieves the best performance when the TTE is 2–3 seconds. On the $\text{JAAD}_{beh}$ dataset, MFT achieves accuracy improvements of 0.15 and 0.02 over Global PCPA [4] and LSOP-Net [23], respectively. On $\text{JAAD}_{all}$, the gains are 0.09 and 0.05, while on the PIE dataset, MFT surpasses them by 0.07 and 0.02. Furthermore, MFT consistently outperforms both baselines across additional evaluation metrics, including AUC, F1 score, and precision, demonstrating its superior robustness and generalization capability. This advantage can be attributed to its use of compact and explicitly semantic contextual attributes. Unlike the implicit and highly entangled nature of raw modality data, these attributes provide explicit and interpretable representations, thereby enhancing robustness and enabling more effective generalization over extended prediction horizons.

**Table 5.** Effect of longer prediction horizon on JAAD$_{beh}$ Dataset

| Models | Time-to-Event | | | | | | | | | |
|---|---|---|---|---|---|---|---|---|---|---|
| | 1-2s | | | | | 2-3s | | | | |
| | Acc | AUC | F1 | Precsion | Recall | Acc | AUC | F1 | Precsion | Recall |
| Global PCPA [4] | 0.62 | 0.54 | 0.74 | 0.65 | 0.85 | 0.53 | 0.47 | 0.65 | 0.62 | 0.68 |
| LSOP-Net [24] | 0.67 | 0.58 | 0.78 | 0.67 | **0.92** | 0.66 | 0.56 | **0.77** | 0.68 | **0.90** |
| MFT | **0.74** | **0.70** | **0.80** | **0.76** | 0.86 | **0.68** | **0.61** | **0.77** | **0.71** | 0.83 |

**Table 6.** Effect of longer prediction horizon on JAAD$_{all}$ Dataset

| Models | Time-to-Event | | | | | | | | | |
|---|---|---|---|---|---|---|---|---|---|---|
| | 1-2s | | | | | 2-3s | | | | |
| | Acc | AUC | F1 | Precsion | Recall | Acc | AUC | F1 | Precsion | Recall |
| Global PCPA [4] | 0.83 | 0.82 | 0.63 | 0.51 | 0.81 | 0.79 | 0.78 | 0.57 | 0.46 | **0.76** |
| LSOP-Net [24] | 0.84 | 0.72 | 0.53 | 0.54 | 0.52 | 0.83 | 0.72 | 0.54 | 0.54 | 0.54 |
| MFT | **0.93** | **0.96** | **0.84** | **0.73** | **0.99** | **0.88** | **0.94** | **0.69** | **0.68** | 0.69 |

**Table 7.** Effect of longer prediction horizon on PIE Dataset

| Models | Time-to-Event | | | | | | | | | |
|---|---|---|---|---|---|---|---|---|---|---|
| | 1-2s | | | | | 2-3s | | | | |
| | Acc | AUC | F1 | Precsion | Recall | Acc | AUC | F1 | Precsion | Recall |
| Global PCPA [4] | 0.89 | 0.86 | 0.80 | 0.79 | 0.81 | 0.78 | 0.77 | 0.65 | 0.59 | 0.73 |
| LSOP-Net [24] | 0.89 | 0.86 | 0.80 | 0.80 | 0.80 | 0.83 | 0.78 | 0.68 | 0.69 | 0.68 |
| MFT | **0.90** | **0.89** | **0.82** | **0.83** | **0.81** | **0.85** | **0.82** | **0.74** | **0.72** | **0.76** |

## 6. Conclusion

In this paper, we propose a novel multi-context fusion Transformer (MFT) for pedestrian crossing intention prediction in urban environments. MFT employs semantically explicit numerical contextual attributes to represent the environment in a lightweight and scalable manner, covering four critical dimensions: pedestrian behavior, environmental conditions, pedestrian localization, and vehicle motion dynamics. It further adopts a progressive fusion strategy, where mutual intra-context and cross-context attention facilitate bidirectional information exchange to capture context-specific and multi-context representations, while guided intra-context and cross-context attention refine the corresponding context tokens and the global CLS token through directed information aggregation. Experimental results demonstrate that MFT achieves competitive performance compared with state-of-the-art methods, attaining accuracies of 74%, 93%, and 90% on JAAD$_{beh}$, JAAD$_{all}$, and PIE datasets, respectively. Ablation studies further demonstrate that different context inputs offer complementary information, collectively improving the model's performance. Moreover, compared with the baseline methods, MFT remains lightweight, requiring the fewest parameters at only 0.95 million and just 23.20 ms for inference.

Future work will mainly focus on lightweight model design and efficient deployment, with additional efforts directed toward validating the proposed network in real-world vehicle and cooperative ITS environments using semantic attributes generated by upstream perception modules, as well as under diverse cultural backgrounds. In addition, we will explore the scalability of the proposed framework to more fine-grained intention prediction, extending the current binary classification setting to multiple intention categories, such as immediate crossing, hesitant crossing, and no crossing.

## Appendix

**Table A1.** Category definitions and encoding scheme for the pedestrian behavior context in the JAAD dataset.

| Attribute | Category | Encoding |
|---|---|---|
| Motion State | Standing | 0 |
| | Walking | 1 |
| Gaze State | Not Looking | 0 |
| | Looking | 1 |
| Head Nod | None | 0 |
| | Nodding | 1 |
| Hand Gesture | Undefined | 0 |
| | Greeting | 1 |
| | Yielding | 2 |
| | Right of Way | 3 |
| | Others | 4 |
| Motion Direction | N/A | 0 |
| | Lateral | 1 |
| | Longitudinal | 2 |

**Table A2.** Category definitions and encoding scheme for the pedestrian behavior context in the PIE dataset.

| Attribute | Category | Encoding |
|---|---|---|
| Motion State | Standing | 0 |
| | Walking | 1 |
| Gaze State | Not Looking | 0 |
| | Looking | 1 |
| Hand Gesture | Undefined | 0 |
| | Greeting | 1 |
| | Yielding | 2 |
| | Right of Way | 3 |
| | Head Nod | 4 |
| | Others | 5 |
| Motion Direction | N/A | 0 |
| | Lateral | 1 |
| | Longitudinal | 2 |

**Table A3.** Category definitions and encoding scheme for the environmental context in the JAAD dataset.

| Attribute | Category | Encoding |
|---|---|---|
| Number of Lanes | Single Lane | 1 |
| | Two Lanes | 2 |
| | Three Lanes | 3 |
| | Four Lanes | 4 |
| Intersection Existence | No | 0 |
| | Yes | 1 |
| Crosswalk Availability | No | 0 |
| | Yes | 1 |
| Traffic Light Status | N/A | 0 |
| | Red | 1 |
| | Green | 2 |
| Traffic Direction | One-way | 0 |
| | Two-way | 1 |
| Road Type | Street | 0 |
| | Parking Lot | 1 |
| | Garage | 2 |
| Stop Sign Presence | Absent | 0 |
| | Present | 1 |
| Pedestrian-related Sign Presence | Absent | 0 |
| | Present | 1 |

**Table A4.** Category definitions and encoding scheme for the environmental context in the PIE dataset.

| Attribute | Category | Encoding |
|---|---|---|
| Number of Lanes | Single Lane | 1 |
| | Two Lanes | 2 |
| | Three Lanes | 3 |
| | Four Lanes | 4 |
| Intersection Type | Midblock | 0 |
| | T-intersection | 1 |
| | T-intersection (left) | 2 |
| | T-intersection (right) | 3 |
| | Four-way Intersection | 4 |
| Crosswalk Availability | No | 0 |
| | Yes | 1 |
| Traffic Light Type | Regular | 0 |
| | Transit | 1 |
| | Pedestrian | 2 |
| Traffic Direction | One-way | 0 |
| | Two-way | 1 |
| Stop Sign Presence | No | 0 |
| | Yes | 1 |
| Pedestrian-related Sign Presence | Absent | 0 |
| | Present | 1 |